\documentclass[10pt,twocolumn,letterpaper]{article}

\usepackage{cvpr}
\usepackage{times}
\usepackage{epsfig}
\usepackage{graphicx}
\usepackage{amsmath}
\usepackage{amssymb}
\usepackage{soul}

\usepackage[pagebackref=true,breaklinks=true,letterpaper=true,colorlinks,bookmarks=false]{hyperref}

\cvprfinalcopy 

\def\cvprPaperID{668} 

\ifcvprfinal\fi

\newcommand{\mypara}{\vspace*{-3mm}\paragraph}
\begin{document}


\title{3DMatch: Learning Local Geometric Descriptors from RGB-D Reconstructions}

\author{
Andy Zeng$^{1}$~~Shuran Song$^{1}$~~Matthias Nie{\ss}ner$^{2}$~~Matthew Fisher$^{2,4}$~~Jianxiong Xiao$^{3}$~~Thomas Funkhouser$^{1}$ \vspace{0.1cm} \\ 
$^{1}$Princeton University\quad\quad\quad\quad
$^{2}$Stanford University\quad\quad\quad\quad $^{3}$AutoX \quad\quad\quad\quad $^{4}$Adobe Systems \\
\href{http://3dmatch.cs.princeton.edu/}{http://3dmatch.cs.princeton.edu\quad\quad\quad}
}

\maketitle



\begin{abstract}
Matching local geometric features on real-world depth images is a challenging task due to the noisy, low-resolution, and incomplete nature of 3D scan data. These difficulties limit the performance of current state-of-art methods, which are typically based on histograms over geometric properties. In this paper, we present 3DMatch, a data-driven model that learns a local volumetric patch descriptor for establishing correspondences between partial 3D data. To amass training data for our model, we propose a self-supervised feature learning method that leverages the millions of correspondence labels found in existing RGB-D reconstructions. Experiments show that our descriptor is not only able to match local geometry in new scenes for reconstruction, but also generalize to different tasks and spatial scales (\eg instance-level object model alignment for the Amazon Picking Challenge, and mesh surface correspondence). Results show that 3DMatch consistently outperforms other state-of-the-art approaches by a significant margin. Code, data, benchmarks, and pre-trained models are available online at \href{http://3dmatch.cs.princeton.edu/}{http://3dmatch.cs.princeton.edu}.

\end{abstract}

\section{Introduction}

\begin{figure}[t]

\includegraphics[width=1\linewidth]{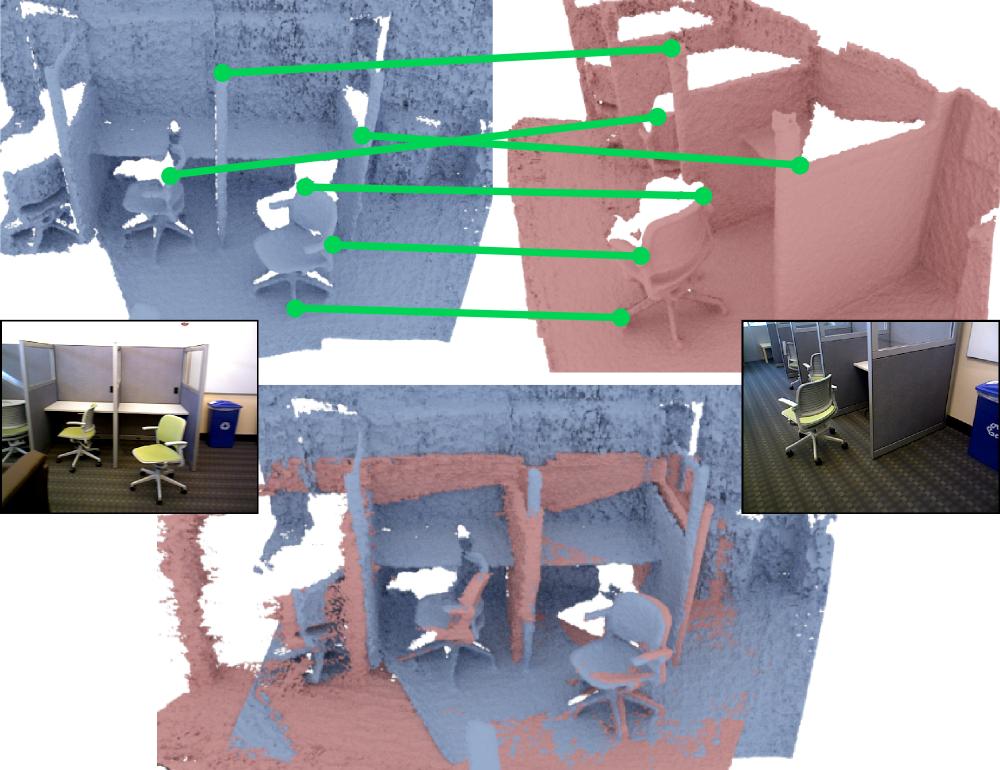}
\caption{In this work, we present a data-driven local descriptor 3DMatch that establishes correspondences (green) to match geometric features in noisy and partial 3D scanning data. This figure illustrates an example of bringing two RGB-D scans into alignment using 3DMatch on depth information only. Color images are for visualization only.}
\label{fig:teaser}
\end{figure}

Matching 3D geometry has a long history starting in the early days of computer graphics and vision. With the rise of commodity range sensing technologies, this research has become paramount to many applications including object pose estimation, object retrieval, 3D reconstruction, and camera localization. 

However, matching local geometric features in low-resolution, noisy, and partial 3D data is still a challenging task as shown in Fig. \ref{fig:teaser}. While there is a wide range of low-level hand-crafted geometric feature descriptors that can be used for this task, they are mostly based on signatures derived from histograms over static geometric properties \cite{johnson1999using,lazebnik2004semi,rusu2009fast}. They work well for 3D models with complete surfaces, but are often unstable or inconsistent in real-world partial surfaces from 3D scanning data and difficult to adapt to new datasets.
As a result, state-of-the-art 3D reconstruction methods using these descriptors for matching geometry require significant algorithmic effort to handle outliers and establish global correspondences \cite{choi2015robust}.

\begin{figure*}
\vspace{-5mm}
\includegraphics[width=1\linewidth]{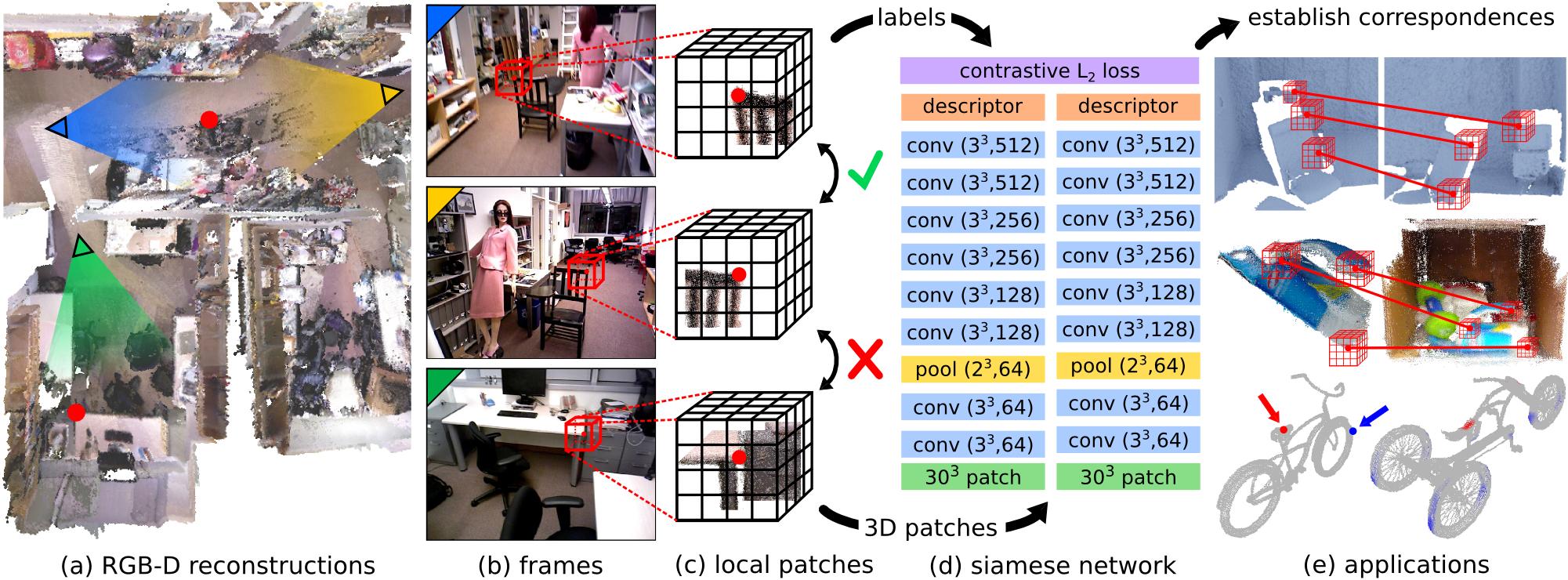}
\caption{{\bf Learning 3DMatch from reconstructions.} From existing RGB-D reconstructions (a), we extract local 3D patches and correspondence labels from scans of different views (b). We collect pairs of matching and non-matching local 3D patches and convert into a volumetric representation (c) to train a 3D ConvNet-based descriptor (d). This geometric descriptor can be used to establish correspondences for matching 3D geometry in various applications (e) such as reconstruction, model alignment, and surface correspondence.
}
\label{fig:overview}
\vspace{-2mm}
\end{figure*}

In response to these difficulties, and inspired by the recent success of neural networks, we formulate a data-driven method to learn a local geometric  descriptor for establishing correspondences between partial 3D data. The idea is that by learning from example, data-driven models can sufficiently address the difficulties of establishing correspondences between partial surfaces in 3D scanning data. To this end, we present a 3D convolutional neural network (ConvNet), called 3DMatch, that takes in the local volumetric region (or 3D patch) around an arbitrary interest point on a 3D surface and computes a feature descriptor for that point, where a smaller distance between two descriptors indicates a higher likelihood of correspondence. 

However, optimizing a 3D ConvNet-based descriptor for this task requires massive amounts of training data (i.e., ground truth matches between local 3D patches). Obtaining this training data with manual annotations is a challenging endeavor. Unlike 2D image labels, which can be effectively crowd-sourced or parsed from the web, acquiring ground truth correspondences by manually clicking keypoint pairs on 3D partial data is not only time consuming but also prone to errors.

Our key idea is to amass training data by leveraging correspondence labels found in existing RGB-D scene reconstructions. Due to the importance of 3D reconstructions, there has been much research on designing algorithms and systems that can build high-fidelity reconstructions from RGB-D data \cite{newcombe2011kinectfusion,niessner2013real,dai2016bundle}.
Although these reconstructions have been used for high-level reasoning about the environment \cite{3DShapeNets,SUN3D}, it is often overlooked that they can also serve as a massive source of labeled correspondences between surfaces points of aligned frames. 
By training on correspondences from multiple existing RGB-D reconstruction datasets, each with its own properties of sensor noise, occlusion patterns, variance of geometric structures, and variety of camera viewpoints, we can optimize 3DMatch to generalize and robustly match local geometry in real-world partial 3D data.

In this paper, we train 3DMatch over 8 million correspondences from a collection of 62 RGB-D scene reconstructions \cite{valentin2016learning,shotton2013scene,SUN3D,lai2014unsupervised,halber2016structured} and demonstrate its ability to match 3D data in several applications. Results show that 3DMatch is considerably better than state-of-the-art methods at matching keypoints, and outperforms other algorithms for geometric registration when combined with standard RANSAC. Furthermore, we demonstrate that 3DMatch can also generalize to different tasks and spatial resolutions.
For example, we utilize 3DMatch to obtain instance-level model alignments for 6D object pose estimation as well as to find surface correspondences in 3D meshes. 
To facilitate further research in the area of 3D keypoint matching and geometric registration, we provide a correspondence matching benchmark as well as a surface registration benchmark similar to \cite{choi2015robust}, but with real-world scan data.


\section{Related Work}

Learning local geometric descriptors for matching 3D data lies at the intersection of computer vision and graphics. We briefly review the related work in both domains.

\mypara{Hand-crafted 3D Local Descriptors.}
Many geometric descriptors have been proposed including Spin Images \cite{johnson1999using}, Geometry Histograms \cite{frome2004recognizing}, and Signatures of Histograms \cite{tombari2010unique}, Feature Histograms \cite{rusu2008aligning}. 
Many of these descriptors are now available in the Point Cloud Library \cite{aldoma2012point}.
While these methods have made significant progress, they still struggle to handle noisy, low-resolution, and incomplete real-world data from commodity range sensors.
Furthermore, since they are manually designed for specific applications or 3D data types, it is often difficult for them to generalize to new data modalities. 
The goal of our work is to provide a new local 3D descriptor that directly learns from data to provide more robust and accurate geometric feature matching results in a variety of settings.

\mypara{Learned 2D Local Descriptors.}
The recent availability of large-scale labeled image data has opened up new opportunities to use data-driven approaches for designing 2D local image patch descriptors.
For instance, various works \cite{simonyan2014learning,simo2015discriminative,yi2016lift,han2015matchnet,vzbontar2014computing,han2015matchnet} learn non-linear mappings from local image patches to feature descriptors.
Many of these prior works are trained on data generated from multi-view stereo datasets \cite{brown2011discriminative}. However, in addition to being limited to 2D correspondences on images, multi-view stereo is difficult to scale up in practice and is prone to error from missing correspondences on textureless or non-Lambertian surfaces, so it is not suitable for learning a 3D surface descriptor. A more recent work \cite{schmidt2017self} uses RGB-D reconstructions to train a 2D descriptor, while we train a 3D geometric descriptor.

\mypara{Learned 3D Global Descriptors.}
There has also been rapid progress in learning geometric representations on 3D data. 
3D ShapeNets \cite{3DShapeNets} introduced 3D deep learning for modeling 3D shapes, and several recent works \cite{maturana_icra_2014,fang20153d,song2015deep} also compute deep features from 3D data for the task of object retrieval and classification. 
While these works are inspiring, their focus is centered on extracting features from complete 3D object models at a \textit{global} level. In contrast, our descriptor focuses on learning geometric features for real-world RGB-D scanning data at a \textit{local} level, to provide more robustness when dealing with partial data suffering from various occlusion patterns and viewpoint differences.

\mypara{Learned 3D Local Descriptors.}
More closely related to this work is Guo \etal \cite{guo20153d}, which uses a 2D ConvNet descriptor to match local geometric features for mesh labeling. However, their approach operates only on synthetic and complete 3D models, while using ConvNets over input patches of concatenated feature vectors that do not have any kind of spatial correlation. In contrast, our work not only tackles the harder problem of matching real-world partial 3D data, but also properly leverages 3D ConvNets on volumetric data in a spatially coherent way.

\mypara{Self-supervised Deep Learning.}
Recently, there has been significant interest in learning powerful deep models using automatically-obtained labels. For example, recent works show that the temporal information from videos can be used as a plentiful source of supervision to learn embeddings that are useful for various tasks \cite{goroshin2015unsupervised,ramanathan2015learning}. Other works show that deep features learned from egomotion supervision perform better than features using class-labels as supervision for many tasks \cite{agrawal2015learning}. Analogous to these recent works in self-supervised learning, our method of extracting training data and correspondence labels from existing RGB-D reconstructions online is fully automatic, and does not require any manual labor or human supervision.

\section{Learning From Reconstructions}
\label{sec:learning-from-reconstructions}
In this paper, our goal is to create a function $\psi$ that maps the local volumetric region (or 3D patch) around a point on a 3D surface to a descriptor vector. Given any two points, an ideal function $\psi$ maps their local 3D patches to two descriptors, where a smaller $\ell_2$ distance between the descriptors indicates a higher likelihood of correspondence.
We learn the function $\psi$ by making use of data from existing high quality RGB-D scene reconstructions. 


The advantage of this approach is threefold: {\bf First}, reconstruction datasets can provide large amounts of training correspondences since each reconstruction contains millions of points that are observed from multiple different scanning views. Each observation pair provides a training example for matching local geometry. Between different observations of the same interest point, its local 3D patches can look very different due to sensor noise, viewpoint variance, and occlusion patterns. This helps to provide a large and diverse correspondence training set. 
{\bf  Second}, reconstructions can leverage domain knowledge such as temporal information and well-engineered global optimization methods, which can facilitate wide baseline registrations (loop closures). We can use the correspondences from these challenging registrations to train a powerful descriptor that can be used for other tasks where the aforementioned domain knowledge is unavailable.
{\bf  Third}, by learning from multiple reconstruction datasets, we can optimize 3DMatch to generalize and robustly match local geometry in real-world partial 3D data under a variety of conditions.  Specifically, we use a total of over 200K  RGB-D images of 62 different scenes collected from Analysis-by-Synthesis~\cite{valentin2016learning}, 7-Scenes~\cite{shotton2013scene}, SUN3D~\cite{SUN3D}, RGB-D Scenes v.2 \cite{lai2014unsupervised}, and Halber \etal~\cite{halber2016structured}. 54 scenes are used for training and 8 scenes for testing. Each of the reconstruction datasets are captured in different environments with different local geometries at varying scales and built with different reconstruction algorithms.

\subsection{Generating Training Correspondences} 
To obtain training 3D patches and their ground truth correspondence labels (match or non-match), we extract local 3D patches from different scanning views around interest points randomly sampled from reconstructions.
To find correspondences for an interest point, we map its 3D position in the reconstruction into all RGB-D frames for which the 3D point lies within the frame's camera view frustum and is not occluded.
The locations of the cameras from which the RGB-D frames are taken are enforced to be at least 1m apart, so that the views between observation pairs are sufficiently wide-baselined.  
We then extract two local 3D patches around the interest point from two of these RGB-D frames, and use them as a matching pair. 
To obtain non-matching pairs, we extract local 3D patches from randomly picked depth frames of two interest points (at least 0.1m apart) randomly sampled from the surface of the reconstruction. 
Each local 3D patch is converted into a volumetric representation as described in Sec. \ref{sec:data-representation}.

Due to perturbations from depth sensor noise and imperfections in reconstruction results, the sampled interest points and their surrounding local 3D patches can experience some minor amounts of drift. We see this jitter as an opportunity for our local descriptor to learn small amounts of translation invariance.
Since we are learning from RGB-D reconstruction datasets using different sensors and algorithms, the jitter is not consistent, which enables the descriptor to generalize and be more robust to it.

\section{Learning A Local Geometric Descriptor}
We use a 3D ConvNet to learn the mapping from a volumetric 3D patch to an 512-dimensional feature representation that serves as the descriptor for that local region.
During training, we optimize this mapping (i.e., updating the weights of the ConvNet) by minimizing the $\ell_2$ distance between descriptors generated from corresponding interest points (matches), and maximize the $\ell_2$ distance between descriptors generated from non-corresponding interest points (non-matches). This is equivalent to training a ConvNet with two streams (i.e., Siamese Style ConvNets \cite{chopra2005learning}) that takes in two local 3D patches and predicts whether or not they correspond to each other.

\subsection{3D Data Representation}
\label{sec:data-representation}
For each interest point, we first extract a 3D volumetric representation for the local region surrounding it. Each 3D region is converted from its original representation (surface mesh, point cloud, or depth map) into a volumetric $30\times{30}\times{30}$ voxel grid of Truncated Distance Function (TDF) values. Analogous to 2D pixel image patches, we refer to these TDF voxel grids as local 3D patches. In our experiments, these local 3D patches spatially span 0.3m\textsuperscript{3}, where voxel size is 0.01m\textsuperscript{3}. The voxel grid is aligned with respect to the camera view. If camera information is unavailable (\ie for pre-scanned 3D models), the voxel grid is aligned to the object coordinates. The TDF value of each voxel indicates the distance between the center of that voxel to the nearest 3D surface. These TDF values are truncated, normalized and then flipped to be between 1 (on surface) and 0 (far from surface). This form of 3D representation is cross-compatible with 3D meshes, point-clouds, and depth maps. Analogous to 2D RGB pixel matrices for color images, 3D TDF voxel grids also provide a natural volumetric encoding of 3D space that is suitable as input to a 3D ConvNet.

\begin{figure}[t]
\vspace{-3mm}
\centering
\includegraphics[width=1\linewidth]{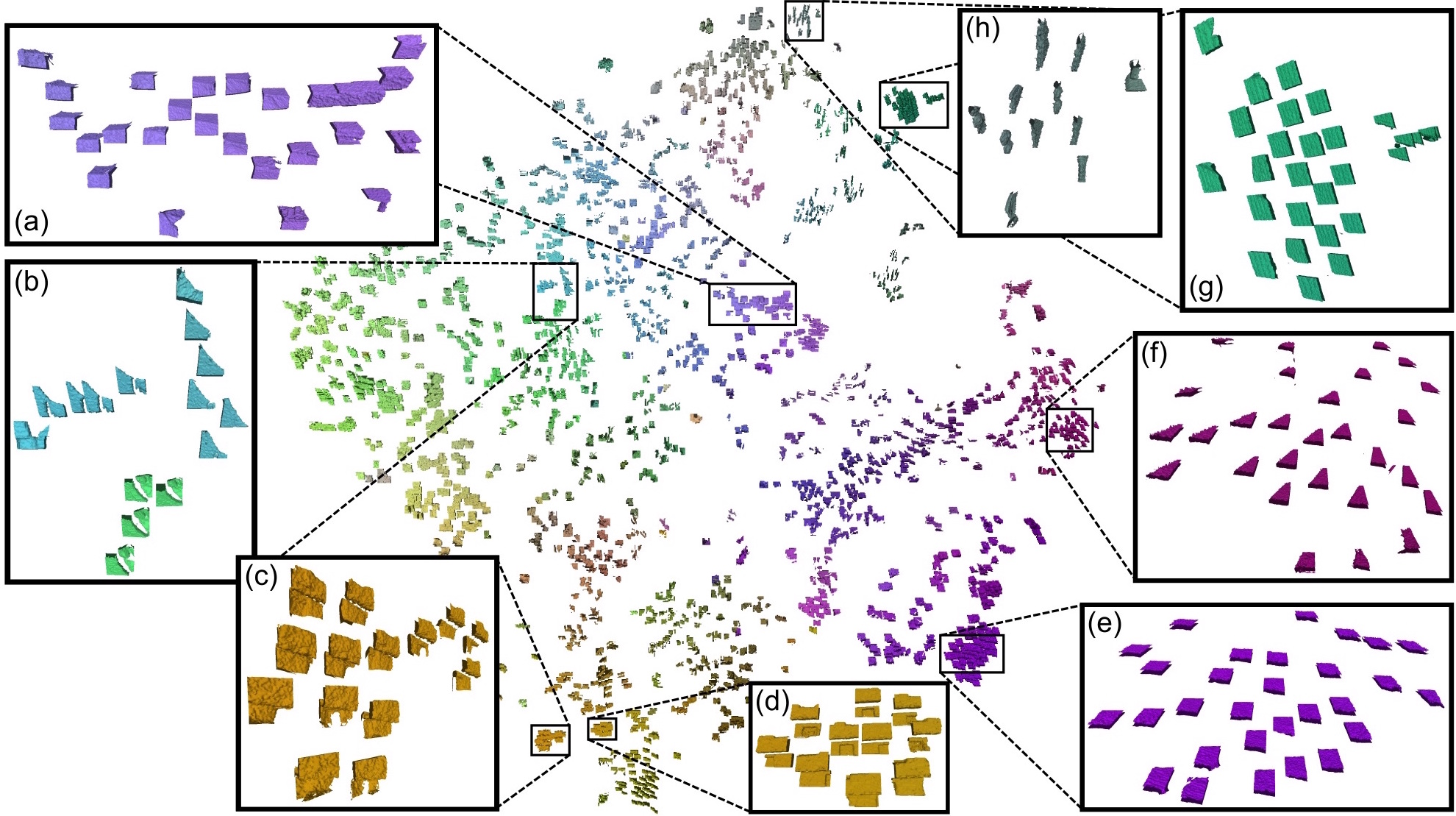}

\caption{{\bf t-SNE embedding of 3DMatch descriptors} for local 3D patches from the RedKitchen test scene of 7-Scenes \cite{newcombe2011kinectfusion}. This embedding suggests that our 3DMatch ConvNet is able to cluster local 3D patches based on local geometric features such as edges (a,f), planes (e), corners (c,d), and other geometric structures (g, b, h) in the face of noisy and partial data.}
\label{fig:tSNE}
\end{figure}

The TDF representation holds several advantages over its signed alternative TSDF \cite{curless1996volumetric},
which encodes occluded space (values near -1) in addition to the surface (values near 0) and free space (values near 1). By removing the sign, the TDF loses the distinction between free space and occluded space, but gains a new property that is crucial to the robustness of our descriptor on partial data: the largest gradients between voxel values are concentrated around the surfaces rather than in the shadow boundaries between free space and occluded space. 
Furthermore, the TDF representation reduces the ambiguity of determining what is occluded space on 3D data where camera view is unavailable.

\subsection{Network Architecture}
3DMatch is a standard 3D ConvNet, inspired by AlexNet \cite{ImageNet}. Given a $30\times{30}\times{30}$ TDF voxel grid of a local 3D patch around an interest point, we use eight convolutional layers (each with a rectified linear unit activation function for non-linearity) and a pooling layer to compute a $512$-dimensional feature representation, which serves as the feature descriptor. Since the dimensions of the initial input voxel grid are small, we only include one layer of pooling to avoid a substantial loss of information. Convolution parameters are shown in Fig.~\ref{fig:overview} as (kernel size, number of filters).

\subsection{Network Training}
During training, our objective is to optimize the local descriptors generated by the ConvNet such that they are similar for 3D patches corresponding to the same point, and dissimilar otherwise. To this end, we train our ConvNet with two streams in a Siamese fashion where each stream independently computes a descriptor for a different local 3D patch. The first stream takes in the local 3D patch around a surface point $p_1$, while the second stream takes in a second local 3D patch around a surface point $p_2$. Both streams share the same architecture and underlying weights.
We use the $\ell_2$ norm as a similarity metric between descriptors, modeled during training with the contrastive loss function \cite{chopra2005learning}.
This loss minimizes the $\ell_2$ distance between descriptors of corresponding 3D point pairs (matches), while pulling apart the $\ell_2$ distance between descriptors of non-corresponding 3D point pairs. During training, we feed the network with a balanced 1:1 ratio of matches to non-matches. This strategy of balancing positives and negatives has shown to be effective for efficiently learning discriminative descriptors \cite{han2015matchnet,simo2015discriminative,yi2016lift}.
Fig.\ref{fig:tSNE} shows a t-SNE embedding \cite{van2008visualizing} of local 3D patches based on their 3DMatch descriptors, which demonstrates the ConvNet's ability to cluster local 3D patches based on their geometric structure as well as local context.

\begin{figure}[t]
\vspace{-3mm}
\centering
\includegraphics[width=1\linewidth]{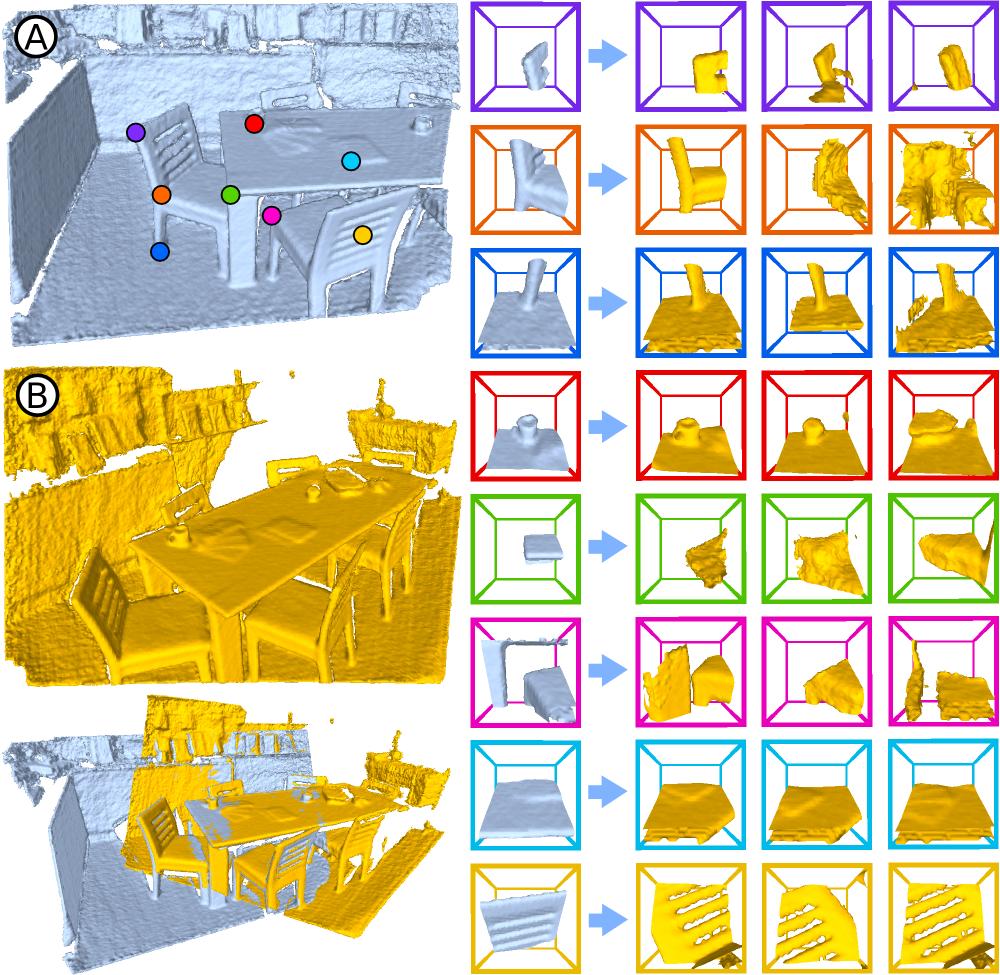}
\caption{{\bf Which 3D patches are matched by 3DMatch?} On the left, we show two fused fragments (A and B) taken at different scan view angles, as well as their registration result using 3DMatch + RANSAC. On the right, each row shows a local 3D patch from fragment A, followed by three nearest neighbor local 3D patches from fragment B found by 3DMatch descriptors. The bounding boxes are color coded to the keypoints illustrated on fragment A.}
\label{fig:patch-matching}
\end{figure}

\section{Evaluation}
In this section, we first evaluate how well our learned local 3D descriptor (3DMatch) can match local 3D patches of interest point pairs (Sec. \ref{sec:keypoint-matching}). We then evaluate its practical use as part of geometric registration for matching 3D data in several applications, such as scene reconstruction (Sec. \ref{sec:geometric-registration}) and 6D object pose estimation (Sec. \ref{sec:generalization-new-domains}).

\subsection{Keypoint Matching}
\label{sec:keypoint-matching}
Our first set of experiments measure the quality of a 3D local descriptor by testing its ability to distinguish between matching and non-matching local 3D patches of keypoint pairs. Using the sampling algorithm described in Sec. \ref{sec:learning-from-reconstructions}, we construct a correspondence benchmark, similar to the Photo Tourism dataset \cite{brown2011discriminative} but with local 3D patches extracted from depth frames. The benchmark contains a collection of $30,000$ 3D patches, with a 1:1 ratio between matches and non-matches. 
As in \cite{brown2011discriminative,han2015matchnet}, our evaluation metric is the false-positive rate (error) at $95\%$ recall, the lower the better.

\mypara{Is our descriptor better than others?}
We compare our descriptor to several other state-of-the-art geometric descriptors on this correspondence benchmark. For Johnson \etal (Spin-Images) \cite{johnson1999using} and Rusu \etal (Fast Point Feature Histograms) \cite{rusu2009fast}, we use the implementation provided in the Point Cloud Library (PCL). While 3DMatch uses local TDF voxel grids computed from only a single depth frame, we run Johnson \etal and Rusu \etal on meshes fused from 50 nearby depth frames to boost their performance on this benchmark, since these algorithms failed to produce reasonable results on single depth frames. 
Nevertheless, 3DMatch outperforms these methods by a significant margin.

\begin{table}[h]
  \centering
  \begin{tabular}{l c c }
    \hline
    Method & Error\\\hline 
    Johnson \etal (Spin-Images) \cite{johnson1999using} & 83.7\\ 
    Rusu \etal (FPFH) \cite{rusu2009fast} & 61.3\\ 
    2D ConvNet on Depth & 38.5\\
    Ours (3DMatch) & \bf{35.3}\\\hline
  \end{tabular}
  \caption{{\bf Keypoint matching task } error (\%) at 95\% recall.\label{table:keypoint-matching}}
\end{table}

\mypara{What's the benefit of 3D volumes vs. 2D depth patches?}
We use TDF voxel grids to represent 3D data, not only because it is an intermediate representation that can be easily converted from meshes or point clouds, but also because this 3D representation allows reasoning over real-world spatial scale and occluded regions, which cannot be directly encoded in 2D depth patches.
To evaluate the advantages of this 3D TDF encoding over 2D depth, we train a variant of our method using a 2D ConvNet on depth patches. The depth patches are extracted from a 0.3m\textsuperscript{3} crop and resized to 64x64 patches. For a fair comparison, the architecture of the 2D ConvNet is similar to our 3D ConvNet with two extra convolution layers to achieve a similar number of parameters as the 3D ConvNet. As shown in Table \ref{table:keypoint-matching}, this 2D ConvNet yields a higher error rate (38.5 vs. 35.3).

\begin{figure*}[t]
\vspace{-5mm}
\centering
\includegraphics[width=1\linewidth]{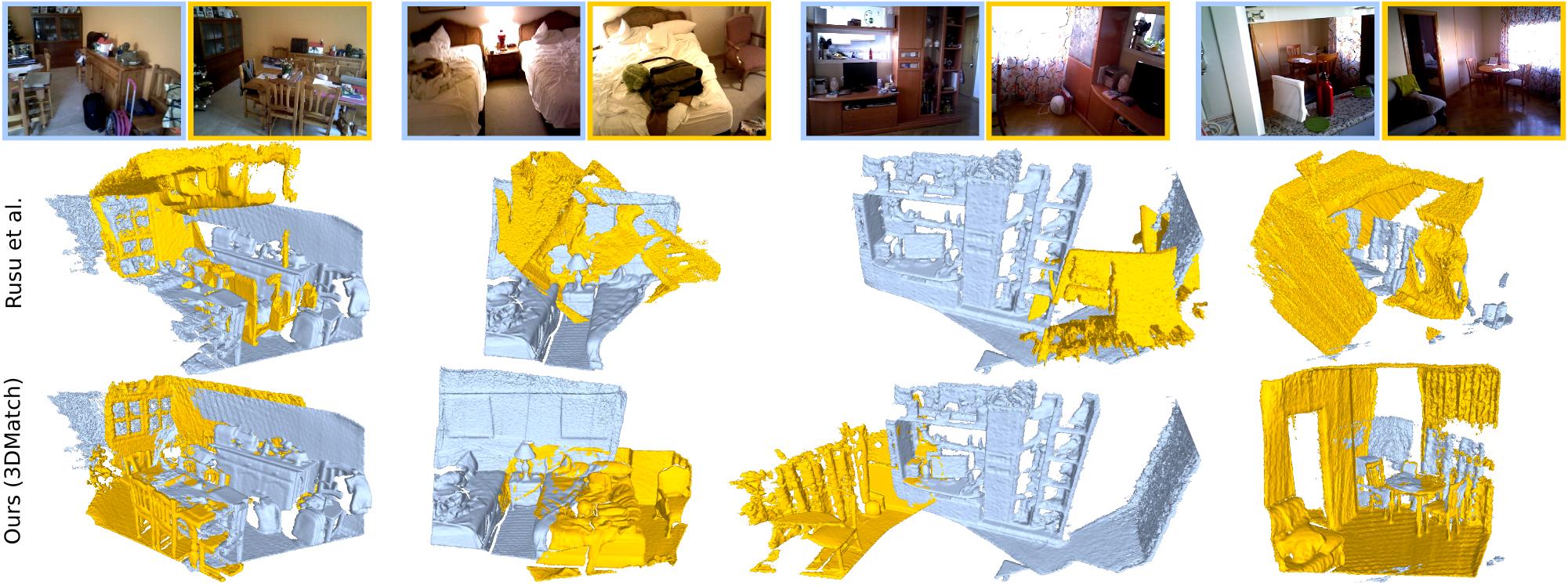}
\caption{{\bf Challenging cases of loop closures from test scenes of SUN3D \cite{SUN3D}}. In these instances, color features in the RGB images (top row) are insufficient for registering the scan pairs due to drastic viewpoint differences. While Rusu \etal \cite{rusu2009fast} fails at aligning the pairs (middle row), 3DMatch is able to successfully align each pair of scans (bottom row) by matching local geometric features.}

\label{fig:loop-closure-example}
\end{figure*}

\mypara{Should we use a metric network?} 
Recent work \cite{han2015matchnet} proposes the joint learning of a descriptor and similarity metric with ConvNets to optimize matching accuracy. To explore this idea, we replace our contrastive loss layer with three fully connected layers, followed by a Softmax layer for binary classification of "match" vs "non-match". We evaluate the performance of this network on our keypoint matching benchmark, where we see an error of 33.1\% (2.2\% improvement).
However, as noted by Yi \etal \cite{yi2016lift}, descriptors that require a learned metric have a limited range of applicability due to the $O(n^2)$ comparison behaviour at test time since they cannot be directly combined with metric-based acceleration structures such as KD-trees. To maintain run-time within practical limits, we use the version of 3DMatch trained with an $\ell_2$ metric in the following sections.

\subsection{Geometric Registration}
\label{sec:geometric-registration}

To evaluate the practical use of our descriptor, we combine 3DMatch with a RANSAC search algorithm for geometric registration, and measure its performance on standard benchmarks. More specifically, given two 3D point clouds from scanning data, we first randomly sample $n$ keypoints from each point cloud. Using the local 3D $30\times{30}\times{30}$ TDF patches around each keypoint (aligned to the camera axes, which may be different per point cloud), we compute 3DMatch descriptors for all $2n$ keypoints. We find keypoints whose descriptors are mutually closest to each other in Euclidean space, and use RANSAC over the 3D positions of these keypoint matches to estimate a rigid transformation between the two point clouds.

\subsubsection{Matching Local Geometry in Scenes}

We evaluate our 3DMatch-based geometric registration algorithm (i.e., 3DMatch + RANSAC) on both real and synthetic datasets. For synthetic, we use the benchmark from Choi \etal \cite{choi2015robust} which contains 207 fragments (each fused from 50 depth frames) of four scenes from the ICL-NUIM dataset \cite{handa2014iclnuim}. However, the duplicated and over-simplified geometry in this ICL-NUIM dataset is very different from that of real-world scenes. Therefore, we create a separate benchmark with fragments formed from the testing split of our real-world reconstruction datasets.
We use the same evaluation scheme introduced by Choi \etal \cite{choi2015robust}, measuring the recall and precision of a method based on two factors: (1) how well it finds loop closures, and (2) how well it estimates rigid transformation matrices. Given two non-consecutive scene fragments $(P_i,P_j)$, the predicted relative rigid transformation $T_{ij}$ is a true positive if (1) over 30\% of $T_{ij}P_i$ overlaps with $P_j$ and if (2) $T_{ij}$ is sufficiently close to the ground-truth transformation $T^{*}_{ij}$. $T_{ij}$ is correct if it brings the RMSE of the ground truth correspondences $K^*_{ij}$ between $P_i$ and $P_j$ below a threshold $\tau = 0.2$
\begin{equation}
\frac{1}{|K^*_{ij}|}\sum_{(p^*,q^*) \in K^*_{ij}}{ ||T_{ij}p^*-q^* ||^2 < \tau^2 }
\label{eq:tau}
\end{equation}
where $p^*$ $q^*$ are the ground truth correspondences.

\begin{figure*}[t]
\vspace{-4mm}
\centering
\includegraphics[width=1\linewidth]{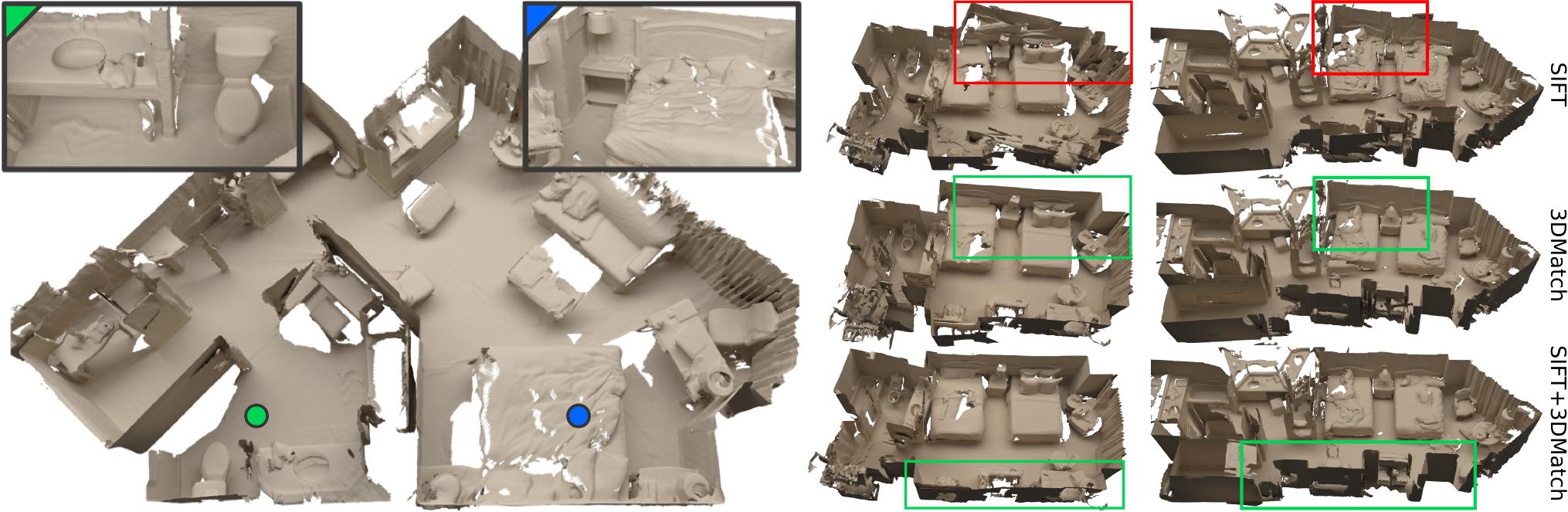}
\caption{{\bf 3DMatch for reconstructions.} On the left, we show a complete reconstruction of an apartment from SUN3D \cite{SUN3D} using only 3DMatch. On the right, we show two reconstructions using only SIFT to match color features (top), using only 3DMatch to match geometric features (middle), and using both SIFT and 3DMatch (bottom). The red boxes highlight areas with poor reconstruction quality, while the green boxes highlight areas with improved quality. These examples show that 3DMatch provides strong geometric feature correspondences that are complementary to color features and can help to improve the quality of reconstructions.}
\label{fig:scene-reconstruction}
\end{figure*}

Since the benchmark from Choi \etal \cite{choi2015robust} uses fragments fused from multiple depth frames, we fine-tune our 3DMatch ConvNet on correspondences over a set of fragments constructed in the same way using the 7-scenes training set. We then run pairwise geometric registration with 3DMatch + RANSAC on every pair of fragments from the benchmarks. We compare the performance of our 3DMatch-based registration approach versus other state-of-the-art geometric registration methods on the synthetic data benchmark in Table \ref{table:geometric-registration-synthetic} \cite{choi2015robust}, and the real data benchmark in Table \ref{table:geometric-registration-real}. We also compare with Rusu \etal \cite{rusu2009fast} and Johnson \etal \cite{johnson1999using} using the same RANSAC-based pipeline. Overall, our descriptor with RANSAC outperforms other methods by a significant margin on both datasets.

\begin{table}[h]
\setlength{\tabcolsep}{2.5 pt}
  \centering
  \begin{tabular}{c c c c c }
    \hline
    Method & Recall (\%) & Precision (\%)\\\hline
    Drost \etal \cite{drost2010model} & 5.3 & 1.6\\
    Mellado \etal \cite{mellado2014super} & 17.8 & 10.4 \\
    Rusu \etal \cite{rusu2009fast} & 44.9 & 14.0 \\
    Choi \etal \cite{choi2015robust} & 59.2 & 19.6 \\
    Zhou \etal \cite{zhou2016fast} & 51.1 & 23.2 \\\hline
    Rusu \etal \cite{rusu2009fast} + RANSAC & 46.1 & 19.1 \\
    Johnson \etal \cite{johnson1999using} + RANSAC & 52.0 & 21.7 \\
    Ours + RANSAC & \bf{65.1} & \bf{25.2} \\\hline
  \end{tabular}
  \caption{\label{table:geometric-registration-synthetic}Performance of geometric registration algorithms between fused fragments of synthetic scans.}
\end{table}

\begin{table}[h]
\setlength{\tabcolsep}{3 pt}
  \centering
  \begin{tabular}{c c c c c }
    \hline
    Method & Recall (\%) & Precision (\%)\\\hline
    Rusu \etal \cite{rusu2009fast} + RANSAC & 44.2 & 30.7 \\
    Johnson \etal \cite{johnson1999using} + RANSAC & 51.8 & 31.6 \\
    Ours + RANSAC & \bf{66.8} & \bf{40.1} \\\hline
  \end{tabular}
  \caption{ \label{table:geometric-registration-real}Performance of geometric registration algorithms between fused fragments of real-world scans.}
\end{table}

\subsubsection{Integrate 3DMatch in Reconstruction Pipeline.}
In this section, we show that 3DMatch is not only capable of detecting challenging cases of loop closure, but also can be used in a standard reconstruction pipeline to generate high-quality reconstructions of new scenes.
We use our 3DMatch descriptor as part of a standard sparse bundle adjustment formulation for scene reconstruction \cite{triggs2000bundle,agarwal2011building}.  Traditionally, sparse RGB features, such as SIFT or SURF, are used to establish feature matches between frames. With 3DMatch, we are able to establish keypoint matches from geometric information and add to the bundle adjustment step. With this simple pipeline we are able to generate globally-consistent alignments in challenging scenes using only geometric information as shown in Fig.~\ref{fig:scene-reconstruction}. 
We also find that color and depth provide complementary information for RGB-D reconstructions. For example, sparse RGB features can provide correspondences where there is insufficient geometric information in the scans, while geometric signals are helpful where there are drastic viewpoint or lighting changes that cause traditional RGB features to fail. 
Fig.~\ref{fig:loop-closure-example} shows challenging cases of loop closure from the testing split of the SUN3D datasets that are difficult for color-based descriptors to find correspondences due to drastic viewpoint differences. Our 3DMatch-based registration algorithm is capable of matching the local geometry to find correspondences and bring the scans into alignment.
In Fig.~\ref{fig:scene-reconstruction}, we show several reconstruction results where combining correspondences from both SIFT (color) and 3DMatch (geometry) improves alignment quality as a whole.


\subsection{Can 3DMatch generalize to new domains?}
\label{sec:generalization-new-domains}

As a final test, we evaluate the ability of our 3DMatch descriptor, which is trained from 3D reconstructions, to generalize to completely different tasks and spatial scales; namely, 6D object pose estimation by model alignment and correspondence labeling for 3D meshes.

\begin{figure}[t]
\vspace{-2mm}
\centering
\includegraphics[width=1\linewidth]{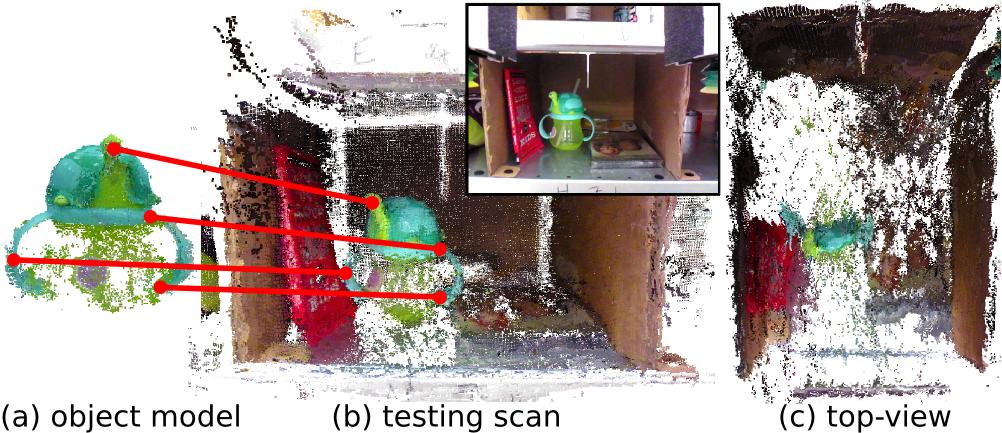}
\caption{\textbf{6D pose estimation in the Amazon Picking Challenge} by aligning object models (a) to scanning data (b). (c) is a top-down view of the scanned shelf highlighting the noisy, partial nature of depth data from the RGB-D sensor.}
\label{fig:apc-matching-problem}
\vspace{2mm}

\centering
\includegraphics[width=1\linewidth]{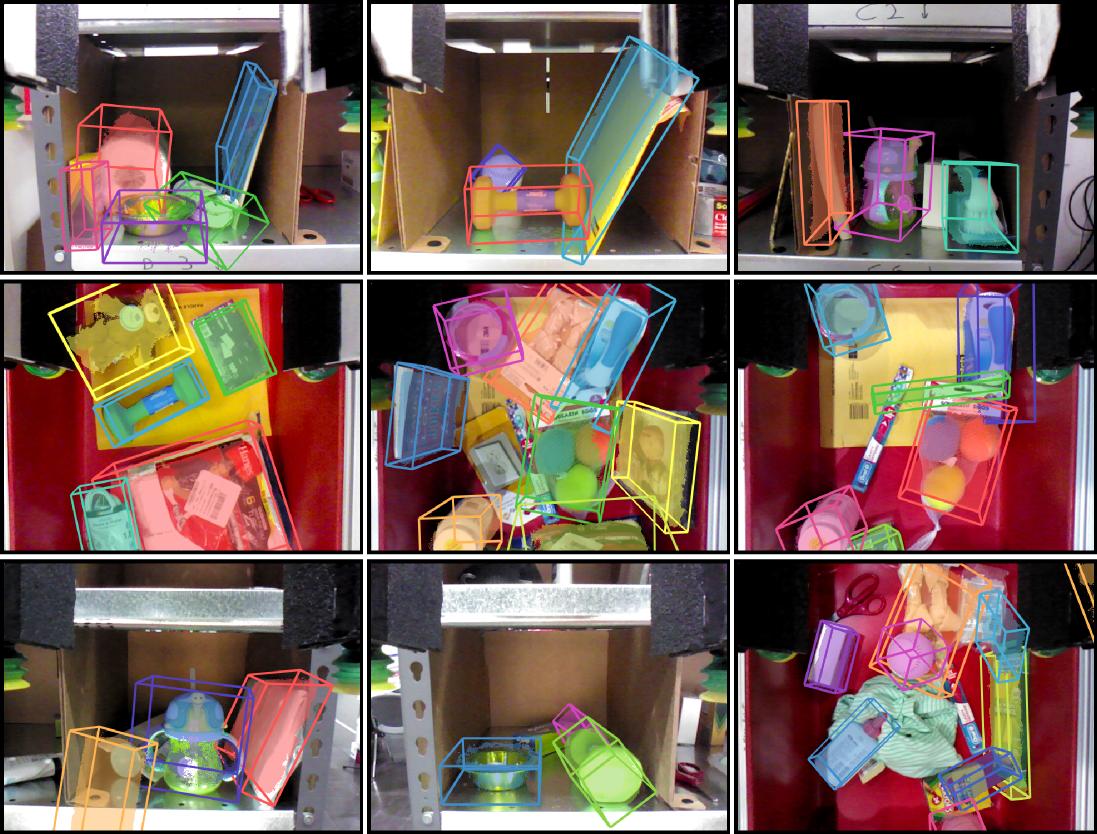}
\caption{\textbf{Predicted object poses on the Shelf \& Tote Benchmark} using 3DMatch + RANSAC. Predicted object poses are shown on the images of the scans with 3D bounding boxes of the transformed object models. 3DMatch + RANSAC works well for many cases; however, it may fail when there is insufficient depth information due to occlusion or clutter (bottom).}
\label{fig:apc-matching-results}
\end{figure}

\mypara{6D Object Pose Estimation by model alignment.}
In our first experiment, the task is to register pre-scanned object models to RGB-D scanning data for the Shelf \& Tote benchmark in the Amazon Picking Challenge (APC) setting \cite{zeng2016multi}, as illustrated in Fig.~\ref{fig:apc-matching-problem}. This scenario is different from scene level reconstruction in the following two aspects: (1) object sizes and their geometric features are much smaller in scale and (2) the alignment here is from full pre-scanned models to partial scan data, instead of partial scans to partial scans.

\begin{figure}[t]
\vspace{-2mm}
\centering
\includegraphics[width=1\linewidth]{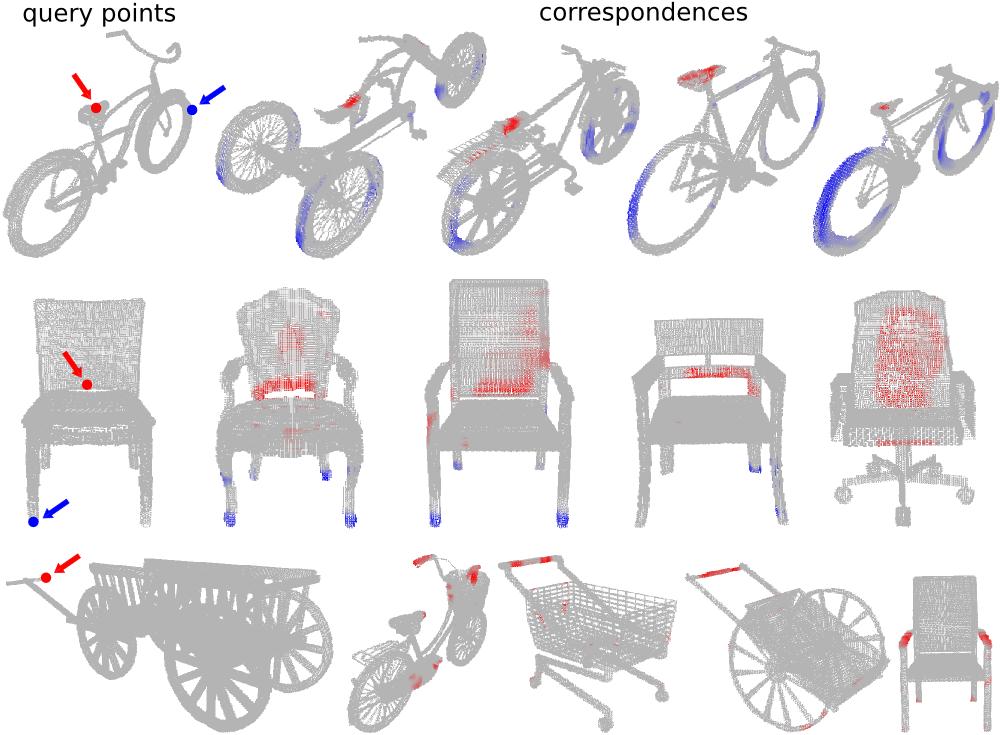}
\caption{\label{fig:mesh-correspondence} {\bf Surface correspondences on 3D meshes.} The first column shows the input mesh and query points (red and blue). The other columns show the respective correspondences found in other meshes of the same object category (top and middle row) and across different object categories (bottom row).}

\end{figure}

\begin{table}[h]
\setlength{\tabcolsep}{0.4 pt}
  \centering
  \begin{tabular}{c c c}
    \hline
    Method & Rotation (\%) & Translation (\%)\\\hline
    Baseline \cite{zeng2016multi} & 49.0 & 67.6 \\
    Johnson \etal \cite{johnson1999using} + RANSAC & 45.5 & 65.9 \\
    Rusu \etal \cite{rusu2009fast} + RANSAC & 43.5 & 65.6 \\
    Ours (no pretrain) + RANSAC & 53.8 & 69.1 \\
    Ours + RANSAC & \bf{61.0} & \bf{71.7} \\\hline
  \end{tabular}
  
  \caption{\label{table:apc-benchmark} Performance of geometric registration algorithms for model-fitting. Numbers are reported in terms of average \% correct rotation and translation predictions.}
\end{table}

To account for spatial scale differences, we reduce the size of each voxel to 0.005m\textsuperscript{3} within the local 3D patches. The voxel grids from the object models are aligned with respect to the object model's coordinates axes, while the voxel grids from the scans are aligned to the camera axes. We use the 3DMatch network pre-trained on reconstructions, and fine-tune it on a 50\% training split of the Shelf \& Tote data. Similar to how we align scene fragments to each other in Sec.~\ref{sec:geometric-registration}, we use a RANSAC based geometric registration approach to align the object models to scans. The predicted rigid transformation is used as object pose. Similar to the baseline approach, we perform model alignment between object models to segmentation results from \cite{zeng2016multi}.

We evaluate on the testing split of the Shelf\&Tote dataset using the error metric from \cite{zeng2016multi}, where we report the percentage of pose predictions with error in orientation smaller than 15$^{\circ}$ and translations smaller than 5cm. We compare to the baseline approach for the Shelf \& Tote benchmark, as well as to other approaches in Table~\ref{table:apc-benchmark}. Several of our predictions are illustrated in Fig.~\ref{fig:apc-matching-results}. Our descriptor significantly outperforms the baseline approach with over 10\% improvement in rotation prediction accuracy and other registration variants. The 3DMatch model without pre-training on reconstructions yields a lower performance, demonstrating the importance of pre-training on reconstruction data.

\mypara{Surface Correspondence on 3D meshes.}
In our final experiment, we test 3DMatch's ability to generalize even further to other modalities. We take a 3DMatch model trained on RGB-D reconstruction data, and directly test it on 3D mesh models without any fine-tuning to see whether 3DMatch is able to find surface correspondences based on local geometry. Given a query point on the surface of a 3D mesh, the goal is to find geometrically similar points on a second mesh (\eg for transferring annotations about human contact points \cite{kim2014shape2pose}). We do this by first encoding the local volumetric regions (with size 0.3m\textsuperscript{3}) of the query point from the first mesh and all surface points from the second mesh into TDF volume aligned to object coordinate, and compute their 3DMatch descriptors. For every surface point on the second mesh, we color it with intensity based on its descriptor's $\ell_2$ distance to the descriptor of the query point. Fig. \ref{fig:mesh-correspondence} shows results on meshes from the Shape2Pose dataset \cite{kim2014shape2pose}. 
The results demonstrate that without any fine-tuning on the mesh data, 3DMatch can be used as a general 3D shape descriptor to find correspondences with similar local geometry between meshes.
Interestingly 3DMatch is also able to find geometric correspondences across different object categories. For example in the third row of Fig. \ref{fig:mesh-correspondence}, 3DMatch is able to match the handles in very different meshes.

\section{Conclusion}
In this work, we presented 3DMatch, a 3D ConvNet-based local geometric descriptor that can be used to match partial 3D data for a variety of applications. 
We demonstrated that by leveraging the vast amounts of correspondences automatically obtained from RGB-D reconstructions, we can train a powerful descriptor that outperforms existing geometric descriptors by a significant margin.
We make all code and pre-trained models available for easy use and integration. To encourage further research, we also provide a correspondence matching benchmark and a surface registration benchmark, both with real-world 3D data.

\mypara{Acknowledgements} 
This work is supported by the NSF/Intel VEC program and Google Faculty Award.
Andy Zeng is supported by the Gordon Y.S. Wu Fellowship.
Shuran Song is supported by the Facebook Fellowship.
Matthias Nie{\ss}ner is a member of the Max Planck Center for Visual Computing and Communications (MPC-VCC).
We gratefully acknowledge the support of NVIDIA and Intel for hardware donations.

\appendix

\section{Appendix}

In this section, we present several statistics of the RGB-D reconstruction datasets used to generate training correspondences for 3DMatch, the implementation details of our network, and run-time statistics relevant to the experiments discussed in the main paper.

\subsection{RGB-D Reconstruction Datasets}
\begin{figure*}
\centering
\includegraphics[width=0.95\linewidth]{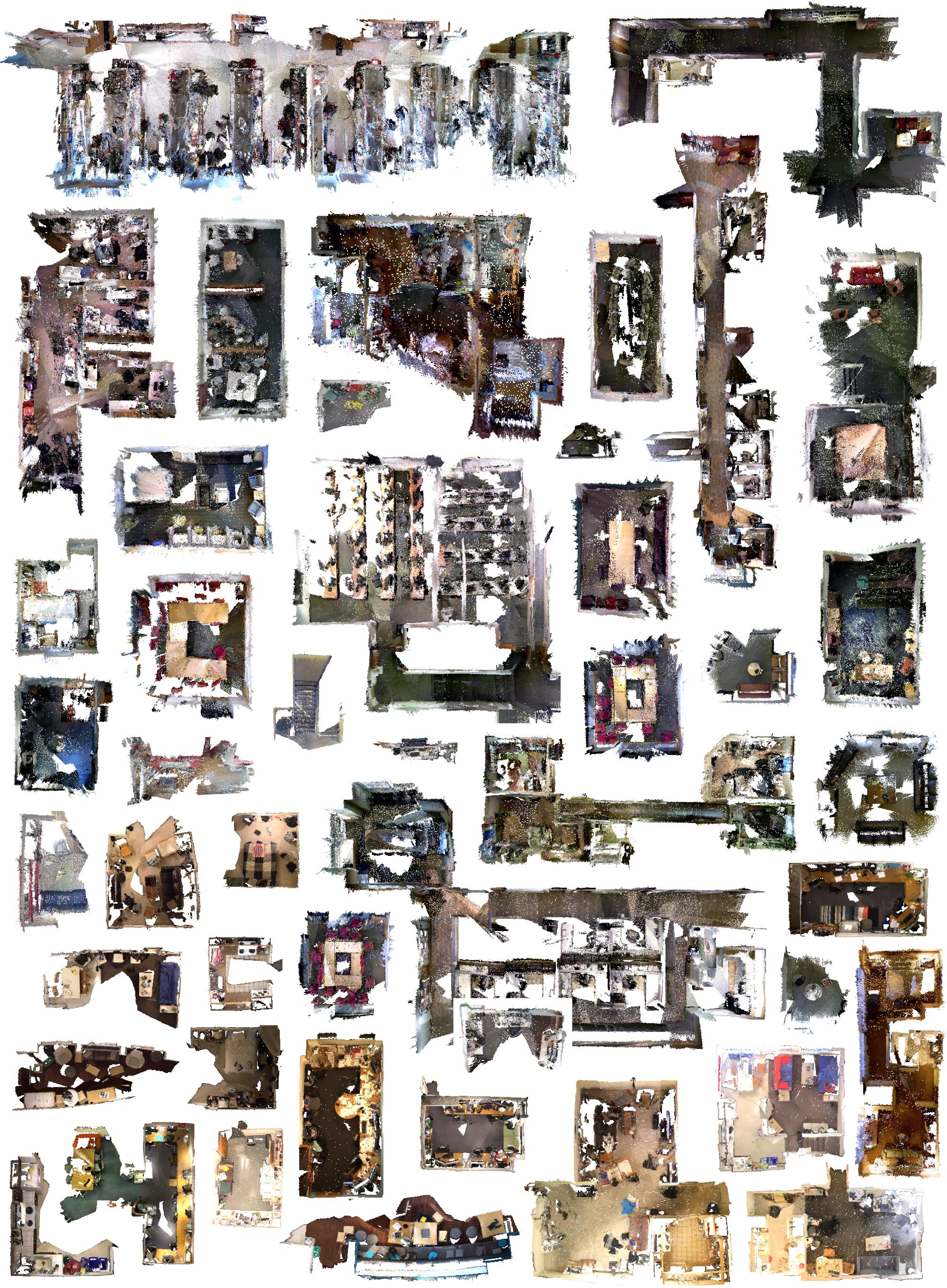}
\caption{Visualizing several RGB-D reconstructions of the datasets used to train 3DMatch in our experiments. Each dataset contains depth scans of different environments with different local geometries at varying scales, registered together by different reconstruction algorithms. These datasets provide a diverse surface correspondence training set with varying levels of sensor noise, viewpoint variance, and occlusion patterns. Color for visualization only.}
\label{fig:reconstruction-datasets}
\end{figure*}

As mentioned in Sec. 3 of the main paper, we use registered depth frames of 62 different real-world scenes collected from Analysis-by-Synthesis~\cite{valentin2016learning}, 7-Scenes~\cite{shotton2013scene}, SUN3D~\cite{SUN3D}, RGB-D Scenes v.2 \cite{lai2014unsupervised}, and Halber \etal~\cite{halber2016structured}, with 54 scenes for training and 8 scenes for testing. For selected scenes of the SUN3D dataset, we use the method from Halber \etal to estimate camera poses. For the precise train/test scene splits, see our project webpage. In Fig. \ref{fig:reconstruction-datasets}, we show top-down views of the completed reconstructions. They are diverse in the environments they capture, with local geometries at varying scales, which provide a diverse surface correspondence training set for 3DMatch. In total, there are 214,569 depth frames over the 54 training scenes, most of which are made up of indoor bedrooms, offices, living rooms, tabletops, and restrooms. The depth sensors used for scanning include the Microsoft Kinect, Structure Sensor, Asus Xtion Pro Live, and Intel RealSense.

\begin{figure*}
\centering
\includegraphics[width=0.95\linewidth]{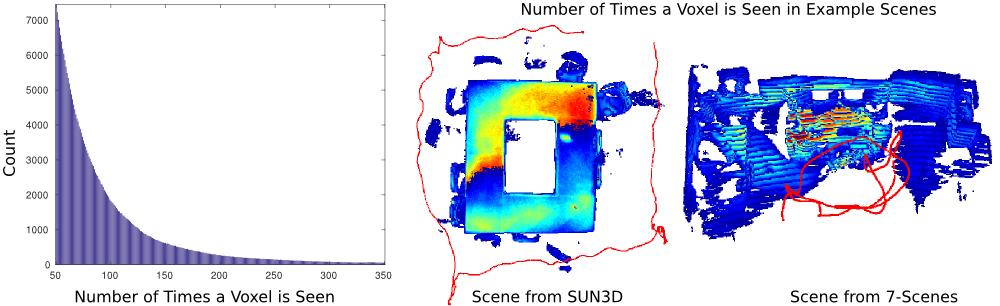}
\caption{Average distribution (by count) of volumetric voxels (size 0.02\textsuperscript{3}m) on the surface vs. the number of frames in which the voxels were seen by the depth sensor in each scene (left). We also illustrate a heat map over the reconstructions of example scenes (right), where a warmer region implies that the surface voxels in area has been seen more times. Regions seen less than 50 times are removed from the illustrations. The camera trajectories are drawn in red.}
\label{fig:dataset-plot}
\end{figure*}

The size of the correspondence training set correlates with the amount of overlap between visible surfaces from different scanning views. In Fig. \ref{fig:dataset-plot}, we show the average distribution of volumetric voxels (size 0.02\textsuperscript{3}m) on the surface vs. the number of frames in which the voxels were seen by the depth sensor. We plot this distribution averaged over the 54 training scenes (left) and illustrate a heat map of over two example reconstructions (right),  where a warmer region implies that the area has been seen more times. The camera trajectories are plotted with a red line.


\subsection{Implementation Details}

We implement the 3DMatch network architecture in Marvin \cite{Marvin20151110}, a lightweight GPU deep learning framework that supports 3D convolutional neural networks. Weights of the network are initialized with the Xavier algorithm \cite{glorot2010understanding}, and biases are initialized to 0. We train by SGD with momentum using a fixed learning rate of 10\textsuperscript{-3}, a momentum of 0.99, and weight decay of 5\textsuperscript{4}. Random sampling matching and non-matching 3D training patches from reconstructions (refer to Sec. 3) is performed on-the-go during network training. We used a batch size of 128, contrastive margin of 1, no batch or patch normalization. Our reference model was trained for approximately eight days on a single NVIDIA Tesla K40c, over 16 million 3D patch pairs, which includes 8 million correspondences and 8 million non-correspondences.


\subsection{Run-time Information}

The following run-times are reported from the implementations used in our experiments. We did not optimize for speed.

\noindent\textbf{TDF Conversion.} As mentioned in Sec 4.1, a local 3D patch around an interest point is represented as a $30 \times 30 \times 30$ voxel grid of TDF values (in all of our experiments, truncation margin is 5 voxels). Converting a point cloud spanning 0.3m\textsuperscript{3} of a depth frame into a TDF voxel grid (using CUDA-enabled GPU acceleration) takes anywhere from 3 - 20 milliseconds, depending on the density of the point cloud. We observe an average conversion time of 8.9 milliseconds per patch from the keypoint matching benchmark in Sec. 5.1 on an NVIDIA Tesla K40c. 


\noindent\textbf{3DMatch Descriptor.} Computing a 3DMatch descriptor (\eg ConvNet forward pass) for a single $30 \times 30 \times 30$ TDF voxel grid takes an average of 3.2 milliseconds with Marvin.

\noindent\textbf{Geometric Registration.} To register two surfaces, we randomly sample $n$ keypoints per surface. $n = 5000$ for scene fragment surface registration (Sec. 5.2.1 - Sec. 5.2.2) and $n = 1000$ for model alignment in the Amazon Picking Challenge (Sec. 5.3). Finding keypoints with mutually closest 3DMatch descriptors takes 2 seconds or less, while RANSAC for estimating rigid transformation can take anywhere from 2 seconds up to a minute depending on the number of RANSAC iterations and rate of convergence. These numbers are reported using a single CPU thread on an Intel Core i7-3770K clocked at 3.5 GHz.

{\small
\bibliographystyle{ieee}
\bibliography{egbib}
}

\end{document}